\documentclass[a4paper, 10pt, conference]{cssconf}
\IEEEoverridecommandlockouts    
\usepackage{epsfig} 
\usepackage[linktoc=page]{hyperref}
\usepackage{subcaption}

\title{\LARGE \bf
Deep-learning-based identification of individual motion characteristics from upper-limb trajectories towards disorder stage evaluation}

\author{Tim Sziburis$^{1,2}$, Susanne Blex$^{2,3}$, Tobias Glasmachers$^{1}$, and Ioannis Iossifidis$^{2}$
\thanks{$^{1}$ Institute for Neural Computation (INI), Ruhr University Bochum, Bochum, Germany}
\thanks{$^{2}$ Institute of Computer Science, Ruhr West University of Applied Sciences, Mülheim an der Ruhr, Germany}
\thanks{$^{3}$ Faculty of Physics and Astronomy, Ruhr University Bochum, Germany}
}

\begin{document}
\maketitle
\thispagestyle{empty}
\pagestyle{empty}

\begin{abstract}
The identification of individual movement characteristics sets the foundation for the assessment of personal rehabilitation progress and can provide diagnostic information on levels and stages of movement disorders.
This work presents a preliminary study for differentiating individual motion patterns using a dataset of 3D upper-limb transport trajectories measured in task-space. Identifying individuals by deep time series learning can be a key step to abstracting individual motion properties.
In this study, a classification accuracy of about 95\,\% is reached for a subset of nine, and about 78\,\% for the full set of 31 individuals.
This provides insights into the separability of patient attributes by exerting a simple standardized task to be transferred to portable systems.
\end{abstract}

\section{INTRODUCTION}
The utilization of individual movement characteristics is essential for the separation of varying phases of motor impairments as well as the evaluation of rehabilitation progress. This can personalize treatment protocols and enhance the success of assist-as-needed therapy. As classical clinical rating scales are gradually superseded by quantitative methods which generate extensive datasets, specialized methods are required to analyze them \cite{Gerald}. Movements can be observed and examined on different levels of motor control abstraction using diverse techniques, leading to a variety of motion evaluation protocols as reviewed in \cite{9857881}.

While a variety of related work focuses on activity recognition across different users \cite{10317707}, there are also studies on user classification or identification. In most cases, relatively complex movement structures like full activities of daily living or gait cycles are analyzed \cite{METIER}. These studies usually incorporate a high number of sensors such as for every joint \cite{9621049}. While offering rich possibilities for research, a simpler set-up proves more practical for continuous tracking of the progression of motion disabilities.

To this effect, we propose a standardized transport task protocol with minimal requirements regarding the utilized motion recording system, as only the task-space path of an object transported by the hand is analyzed. A deep time series learning strategy is applied to investigate the separability of individual participants' movement trajectories. 

In this preliminary study, data of participants without known movement disorders are evaluated to demonstrate that the used approach can differentiate between individuals. In future work, it is to be expanded to differentiate between different stages of disorders and rehabilitation, respectively.

\section{MATERIAL AND METHODS}
\begin{figure}
	\centering
	\begin{minipage}{.48\linewidth}
		\centering
		\includegraphics[width=1.055\linewidth]{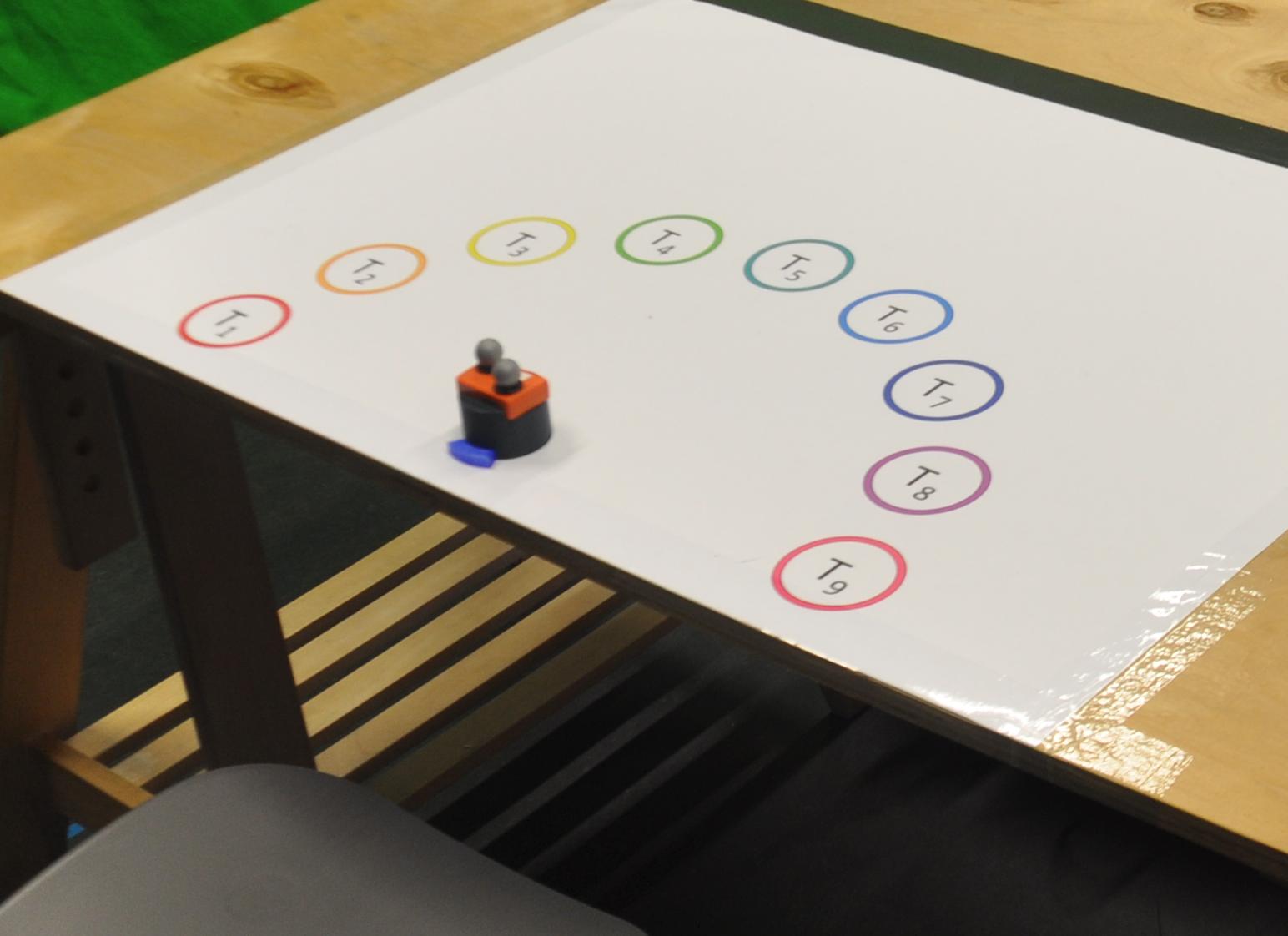}
		\captionof{figure}{Experimental set-up}
		\label{fig:setup}
	\end{minipage}%
	\hfill
	\begin{minipage}{.48\linewidth}
		\centering
		\includegraphics[width=\linewidth]{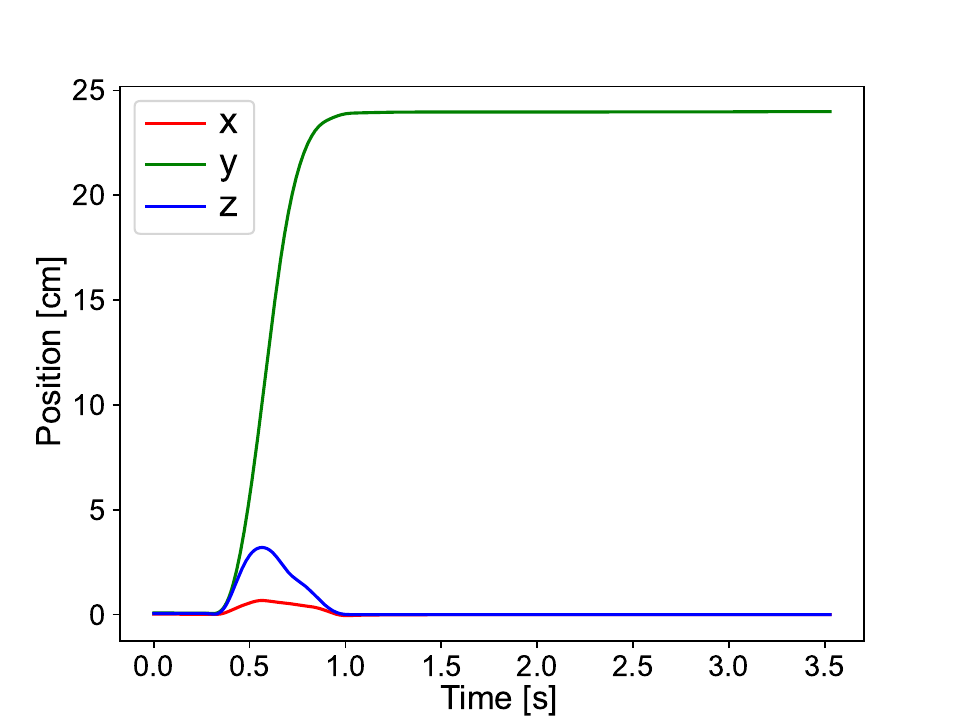}
		\captionof{figure}{Typical trial path}
		\label{fig:trial}
	\end{minipage}
\end{figure}

This work makes use of the Ruhr Hand Motion Catalog \cite{sziburis2023ruhrhandmotioncatalog} which consists of 31 human participants' center-out trajectories captured during three-dimensional movement (age range 21--78). An object is transported from a unified start location to one of nine target positions in randomized order while sitting. The experimental set-up is shown in Fig. \ref{fig:setup}.

In this study, only the positional course from the transported object located in the right hand is considered. A representative example of one trial's positional course over time is depicted in Fig. \ref{fig:trial}. 
The dataset provides trajectory data from two systems measured simultaneously, namely a portable IMU device and an optical motion capture system. In this work, only the latter is used as part of a preliminary study, since it represents the reference data for proof of concept.

The advantage of deep learning compared to classical machine learning is its independence from explicit feature generation. The model used to automatically find patterns for user identification is an adapted ResNet18 architecture for residual learning with the possibility to bypass network layers in the training process, as originally proposed in \cite{resnet} for computer vision. Since then, it has also been shown as competitive for time-series learning \cite{7966039} and constitutes a good choice if computational resources are comparably limited, as in embedded systems or other real-time scenarios.

The trajectory data, i.\,e. the course of the marker's position, were captured at a sampling rate of 250\,Hz and filtered by a 4th-order low-pass Butterworth filter with cut-off at 7\,Hz. Furthermore, windowing with 7 samples and no overlap was applied to the z-score normalized data, before training was conducted in 100 epochs. The data were split into 90\,\% for training and validation (80\,\%/20\,\%) and 10\% for testing, while maintaining a homogeneous distribution of participants across splits. For performance evaluation, a 10-fold cross-validation was applied.

\section{RESULTS}
Two scenarios were examined: the first with 9 randomly chosen participants (equidistantly distributed between 0 and 30), the second with all 31 participants of the dataset.

\begin{figure}[h!]
	\centering
	\includegraphics[width=0.75\linewidth]{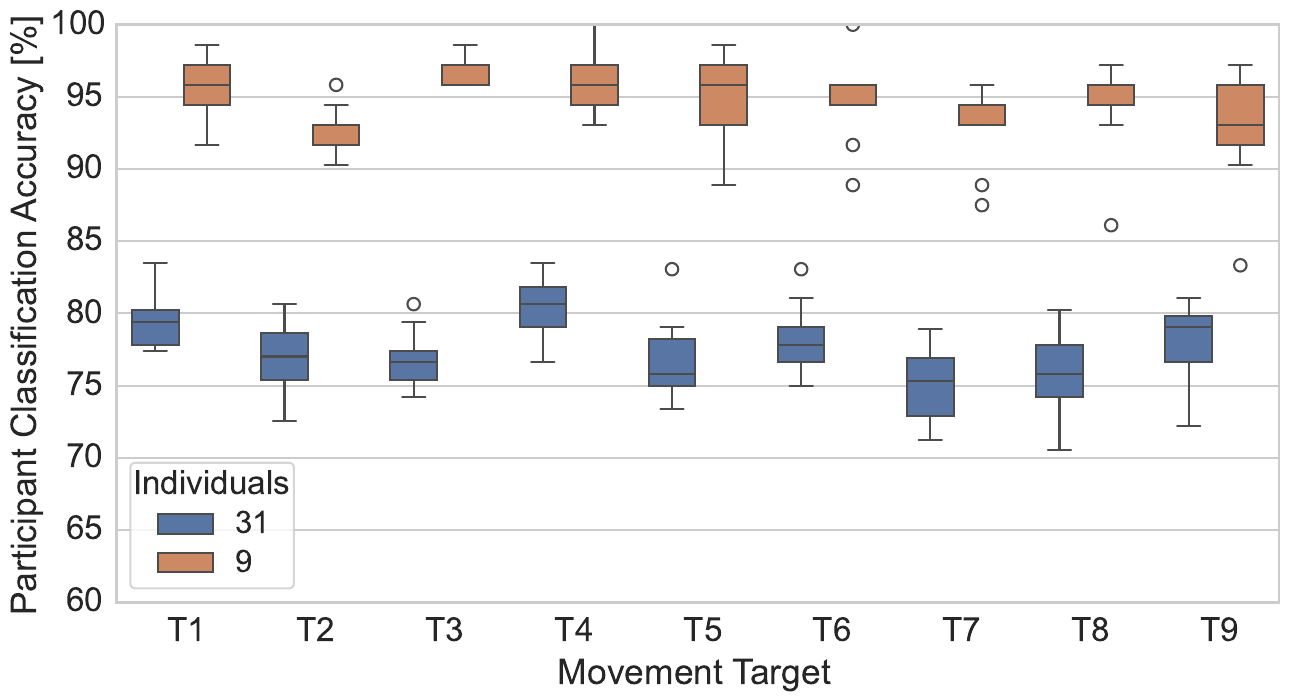}
    \caption{Participant classification performance (10 folds) with respect to task, for scenarios of 9 and 31 participants.}
    \label{fig:res1}
\end{figure}

For 9 participants, an accuracy in classifying persons of about 95\% could be achieved on average. For 31 participants, the mean accuracy was at about 78\,\%. For both scenarios, there was no remarkable difference with respect to the target. However, the highest medians of correct predictions occurred in the movement to targets 1 and 4 (see Fig. \ref{fig:res1}), i.\,e. the most left and the one left to the middle target.

\begin{figure}[h!]
	\centering
	\includegraphics[width=0.8\linewidth]{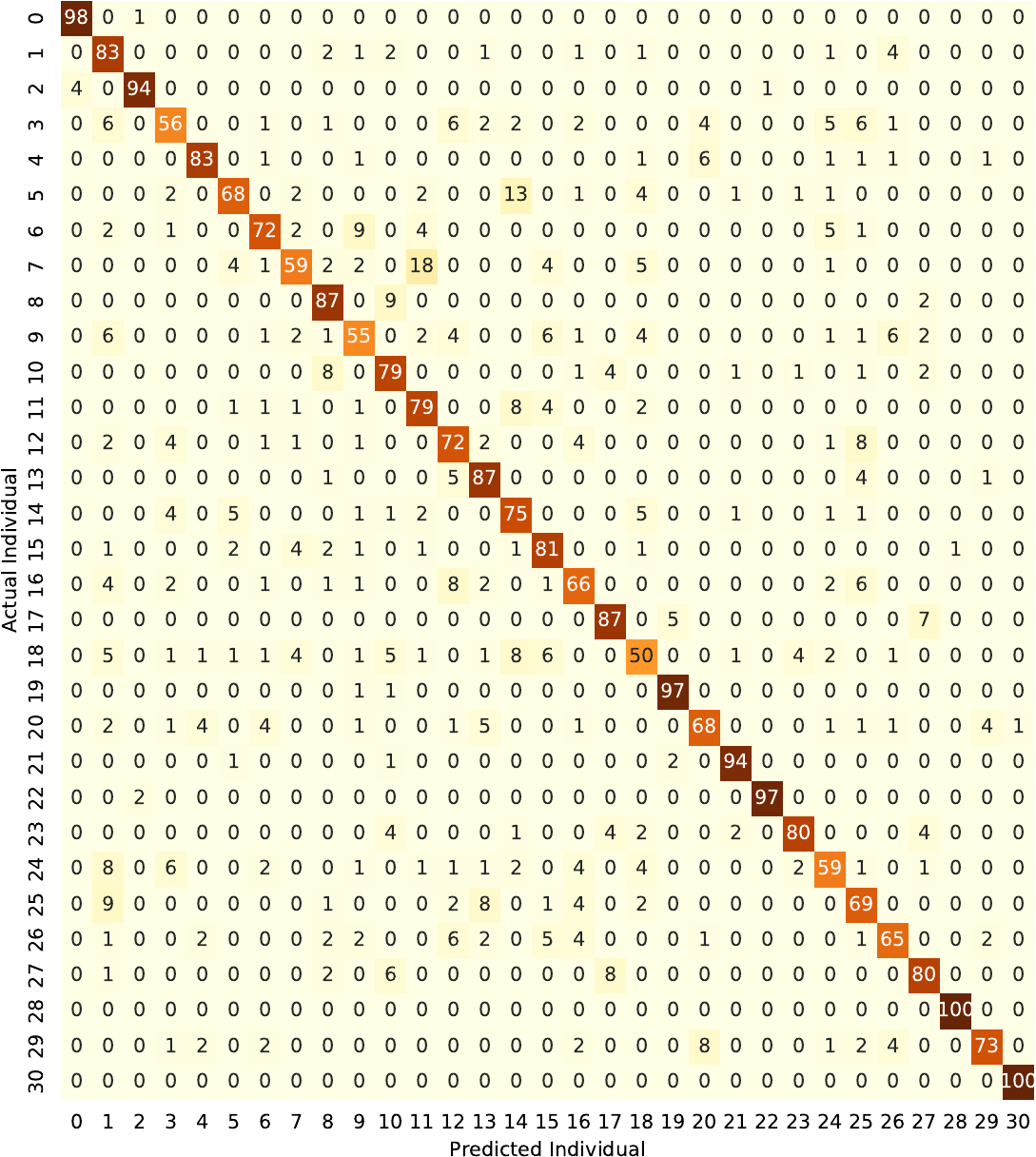}
     \caption{Rounded percentual confusion matrix for correct individual classifications (all 31 participants).}
	\label{fig:confu}
\end{figure}

When examining the confusion matrix for the scenario of all 31 participants (Fig. \ref{fig:confu}), there is a participant-depending variety of 50\% to 100\% correct classifications.

\section{DISCUSSION}
The difference in results between 31 and the subset of 9 participants might be due to effects of movement similarity: An important factor could be age, as the participants were ordered by age and the 9 participants were selected by fixed number intervals from all 31. These larger intervals between the ages may have enhanced accuracy.

While further learning models might improve the identification, the classification accuracy for 31 persons of about 78\% points towards common, non-separable movement patterns and the similarity of participants' movement conditions. Some individuals are only separable to a certain extent, as shown in Fig. \ref{fig:confu}, and form clusters which differentiate as a whole from single perfectly separable individuals. This is desirable for disorder and rehabilitation stage analysis. Otherwise, the separability of participants themselves could counteract the separability of pathological conditions. However, if age is encoded in the movement, this already might signal health-related effects on separability.

Although more targets increase robustness, only one target might be sufficient since the performance is similar for all single tasks. This has to be evaluated on an individual basis.

Due to the low filtering frequency, the approach is transferable to systems with lower sampling rates. While the dataset additionally provides trajectory data from an IMU, the current study used optical motion capture as a reference until the quality of the portable sensor is validated. Then, the proposed methodology can be applied to real-time mobile diagnosis systems if the neural network is trained beforehand. The pre-trained model has to be tested for transferability across users, in consideration of data protection.

\section{CONCLUSIONS}
The preliminary study results show that the presented data-based approach for individual classification is a promising step towards monitoring health conditions and personalizing rehabilitation protocols. It is based on an easily reproducible task, which only requires the 3D course of a transport object's position. The general approach might also be applicable to trajectory data from other movement tasks or differing parts of the body.

Our work will be extended by data from different stages of movement disorders which manifest in upper-limb transport trajectories. Furthermore, we will present a model-based method to provide generic modelling for the movement task and trajectory deviations stemming from motor impairments.

\bibliographystyle{./IEEEtran}
\bibliography{bibliography}

\end{document}